# The impact of tissue detection on diagnostic artificial intelligence algorithms in digital pathology

Sol Erika Boman[1,4], Nita Mulliqi[1], Anders Blilie[2,3], Xiaoyi Ji[1], Kelvin Szolnoky[1], Einar Gudlaugsson[2], Emiel A.M. Janssen[2,18,19], Svein R. Kjosavik[5,6], José Asenjo[7], Marcello Gambacorta[8], Paolo Libretti[8], Marcin Braun[9], Radzislaw Kordek[9], Roman Łowicki[10], Kristina Hotakainen[11,12], Päivi Väre[13], Bodil Ginnerup Pedersen[14,15], Karina Dalsgaard Sørensen[15,16], Benedicte Parm Ulhøi[17], Lars Egevad[20], Kimmo Kartasalo[21]

1. Department of Medical Epidemiology and Biostatistics, Karolinska Institutet, Stockholm, Sweden
2. Department of Pathology, Stavanger University Hospital, Stavanger, Norway
3. Faculty of Health Sciences, University of Stavanger, Stavanger, Norway
4. Department of Molecular Medicine and Surgery, Karolinska Institutet, Stockholm, Sweden
5. The General Practice and Care Coordination Research Group, Stavanger University Hospital, Norway
6. Department of Global Public Health and Primary Care, Faculty of Medicine, University of Bergen, Norway
7. Department of Pathology, Synlab, Madrid, Spain
8. Department of Pathology, Synlab, Brescia, Italy
9. Department of Pathology, Chair of Oncology, Medical University of Lodz, Lodz, Poland
10. 1st Department of Urology, Medical University of Lodz, Lodz, Poland
11. Department of Clinical Chemistry and Hematology, University of Helsinki, Helsinki, Finland
12. Laboratory Services, Mehiläinen Oy, Helsinki, Finland
13. Department of Pathology, Mehiläinen Länsi-Pohja Hospital, Kemi, Finland
14. Department of Radiology, Aarhus University Hospital, Aarhus, Denmark
15. Department of Clinical Medicine, Aarhus University, Aarhus, Denmark
16. Department of Molecular Medicine, Aarhus University Hospital, Aarhus, Denmark
17. Department of Pathology, Aarhus University Hospital, Aarhus, Denmark
18. Department of Chemistry, Bioscience and Environmental Engineering, University of Stavanger, Stavanger, Norway
19. Institute for Biomedicine and Glycomics, Griffith University, Queensland, Australia
20. Department of Oncology and Pathology, Karolinska Institutet, Stockholm, Sweden
21. Department of Medical Epidemiology and Biostatistics, SciLifeLab, Karolinska Institutet, Stockholm, Sweden

Corresponding author: Kimmo Kartasalo, kimmo.kartasalo@ki.se.

# Abstract

Tissue detection is a crucial first step in most digital pathology applications. Details of the segmentation algorithm are rarely reported, and there is a lack of studies investigating the downstream effects of a poor segmentation algorithm. Disregarding tissue detection quality could create a bottleneck for downstream performance and jeopardize patient safety if diagnostically relevant parts of the specimen are excluded from analysis in clinical applications.

This study aims to determine whether performance of downstream tasks is sensitive to the tissue detection method, and to compare performance of classical and AI-based tissue detection. To this end, we trained an AI model for Gleason grading of prostate cancer in whole slide images (WSIs) using two different tissue detection algorithms: thresholding (classical) and UNet++ (AI). A total of 33,823 WSIs scanned on five digital pathology scanners were used to train the tissue detection AI model. The downstream Gleason grading algorithm was trained and tested using 70,524 WSIs from 13 clinical sites scanned on 13 different scanners.

There was a decrease from 116 (0.43%) to 22 (0.08%) fully undetected tissue samples when switching from thresholding-based tissue detection to AI-based, suggesting an AI model may be more reliable than a classical model for avoiding total failures on slides with unusual appearance. On the slides where tissue could be detected by both algorithms, no significant difference in overall Gleason grading performance was observed. However, tissue detection dependent clinically significant variations in AI grading were observed in 3.5% of malignant slides, highlighting the importance of robust tissue detection for optimal clinical performance of diagnostic AI.

# Introduction

Digital pathology involves analyzing histopathological whole slide images (WSIs) using computer methods.[1] Performance of artificial intelligence (AI) within this field has become increasingly accurate, and is now approaching expert pathologist level in a wide variety of tasks. Notably, digital pathology has seen recent successes in the fields of diagnostics,[2–5] such as cancer classification and grading; genomics,[6–11] such as detecting the presence of microsatellite instability or specific genetic mutations; and most recently, prognostics,[12,13] foregoing intermediate diagnostic steps and predicting outcomes directly. With these developments, clinical application of AI systems has become relevant,[14,15] resulting in an increased need for quality and safety of these systems.[16,17]

An early step in a large number of digital pathology tasks involves segmentation of the region of interest (ROI).[3,7–11,13,18–24] By extracting and analysing only the ROI, computation time is heavily reduced while ensuring only the relevant parts of the image are analyzed. In some cases the ROI is a specific part of the sample, such as nuclei, glands, epithelia or lumina, but in the most general case it represents all tissue distinguished from image background. The general problem of segmentation involves categorizing each pixel of an image into different classes or objects. In this paper, "tissue detection" refers to binary tissue segmentation, where all tissue pixels constitute the ROI.

Segmentation tasks typically require manual labeling by a pathologist or the use of AI models, the latter of which requires training labels that may not necessarily need to be of the same quality as manual labeling.[25] Tissue detection, however, is often simple enough to be done automatically using classical image analysis techniques such as thresholding or edge detection, as is common practice.[3,7–10,13,19–21,24,26] However, these methods also rely on user-defined parameters, the optimal choice of which may vary greatly depending on characteristics of the WSIs, such as the scanner used to digitize them. Indeed, generalisation across scanners, labs, and patient populations is crucial even at the tissue detection step. Despite being a frequent processing step, it is not common practice to include segmentation parameters in digital pathology articles, and some papers even omit the tissue detection method entirely.[27,28]

Several open source digital pathology pipelines that include tissue detection have been published,[29–32] sometimes arguing for the standardized use of these algorithms within the field of digital pathology, but these also fail to report tissue detection accuracy, and none of these algorithms have become standard.[33] Comparisons of tissue detection methods exist[34], but we are not aware of investigations into their downstream performance effects in larger AI systems. Because a failure to detect tissue could, in the worst case, lead to exclusion of malignant tissue from analysis, consistent tissue detection is crucial in a clinical context.

Our main hypothesis is that as AI models become increasingly precise, the overall performance of a diagnostic system risks being constrained by the quality of the initial tissue detection step. We tested this hypothesis in the context of Gleason grading of prostate cancer in biopsies, using a state-of-the-art AI model.[35] For the tissue detection step of the system, we compared a classical algorithm using Otsu's thresholding[3,36] and an AI method. We compared the two methods both in terms of directly measuring tissue detection performance and in terms of the resulting downstream performance of the Gleason grading AI which was developed in accordance to a pre-specified study protocol.[37]

# Methods

## Study design

The study had two distinct steps: development and evaluation of an AI tissue detection algorithm, and evaluation of downstream performance of a Gleason grading algorithm, comparing the AI tissue detection with that of a classical thresholding-based tissue detection method. The former was done by procuring an evaluation set of high quality tissue segmentation masks to compare against, then empirically finding a high performing architecture and training set (more details below). The latter utilized an end-to-end Gleason grading model presented recently[35]. A separate study protocol[37] provides details on the development and evaluation of the Gleason grading model, including how reference grading by pathologists was obtained for each cohort.

Prior to conducting this study, we had generated segmentation masks for tissue detection for every WSI using Otsu's thresholding, to be used for development of the Gleason grading algorithm.[35] The parameters of the thresholding algorithm had been selected individually for each cohort. These segmentation masks were used as labels for the training set of the AI tissue detection algorithm, which used a UNet++ architecture.[38] Subsets of the segmentation masks had been checked visually, and in certain cases manually edited to improve quality (see **Table 1**). Utilising some of these, as well as by iteratively re-running the thresholding algorithm with parameters tweaked on a WSI-by-WSI basis until high quality segmentation masks were confirmed by visual inspection, a set of 6,823 WSIs with high quality masks was generated. This set was used both for continuously validating the AI during training, as well as for evaluating its performance after finishing training for the purpose of model selection.

Two sets of segmentation masks were generated for each WSI in the test set of the Gleason grading AI model. One set was generated using thresholding, using a single set of manually fine-tuned parameters for all WSIs. The other set was generated by running the AI model from the previous step, with no additional processing steps. The Gleason grading algorithm was then evaluated twice, once with tissue detection based on each set of segmentation masks. To allow for a direct comparison of downstream task performance, only those WSIs where both segmentation algorithms detected tissue were included. This excluded a few difficult cases where one or both algorithms failed to detect any tissue.

## Datasets and data partitioning

The dataset represents digitized hematoxylin and eosin (H&E) stained prostate core needle biopsies from patients who underwent biopsy between 2012 and 2023. Samples were obtained from 15 clinical sites, of which this study utilized slides from 13, excluding the AQ and KUH-2 cohorts representing non-gradable rare variants (see the study protocol for cohort abbreviations and descriptions). The included slides were scanned using 13 whole slide scanners comprising 9 different models from 5 different vendors. The Gleason grading AI was trained on 55,798 WSIs (STHLM3, SUH) and tuned on 1,177 WSIs (STHLM3, RUMC, KUH-1). For this study, evaluation of the Gleason grading AI was done on 18,848 WSIs (AUH, MLP, MUL, RUMC, SCH, SFI, SFR, SPROB20, STHLM3, SUH, UKK, WNS) from the internal and external validation cohorts. Internal validation cohorts represent data from the same lab and/or WSI scanner as the training data but from independent patients, and external validation cohorts represent data from different labs, scanners, and patients than the training data.[37]

The WSIs for developing the segmentation AI model were selected from the development set of the Gleason grading AI to ensure that the combined system respected the held-out internal and external test set splits specified in the study protocol. Multiple segmentation models were trained using a few different subsets of the data before choosing the UNet++ architecture trained on 33,823 WSIs (Table 1), as this achieved the highest tissue detection sensitivity on the segmentation evaluation set. Of these 33,823 WSIs, 6,172 (18.2%) had strong labels that were either checked visually to verify their quality, or, in 54 cases (0.16% of total), manually edited. The other 27,651 (81.8%) WSIs had weak labels that had not been checked for quality, apart from during the initial mask creation process when parameters were tuned empirically using a small and random subset of the WSIs. Validation for early stopping and evaluation for model selection used the set of 6,823 WSIs with manually curated high-quality segmentation masks (see "Study design"), split 30%-70% on patient-level (2,523 WSIs for validation and 4,305 WSIs for evaluation, see Table 1).

## Thresholding algorithm

The thresholding algorithm used for the comparison, to generate labels for training the AI model, and to generate the validation and evaluation sets, was based on Otsu's method[36] and subsequent morphological operations. Specifically, it applies the following scipy and scikit-image Python functions to WSIs downsampled to a resolution of 8.0 μm per pixel:

> 1. convolution (scipy.signal.convolve2d) using the isotropic "Mehrstellen" nine-point stencil of the two-dimensional Laplacian operator[39] as a kernel,
> 2. Otsu's thresholding (skimage.filters.threshold_otsu) on the convoluted image to find the optimal threshold, creating a binary image using values of the convoluted image greater than the threshold,

3. binary closing (skimage.morpholoy.binary_closing) on the resulting binary image,

4. binary opening (skimage.morphology.binary_opening),

5. filling any small holes (scipy.ndimage.morphology.binary_fill_holes),

6. removing thin objects (skimage.measure.regionprops.minor_axis_length),

7. removing hues, saturations, and values outside a specified range (skimage.color.rgb2hsv),

8. clearing borders (skimage.segmentation.clear_border).

The original purpose of the implementation of this algorithm was to generate segmentation masks to be used for the Gleason grading AI, and these are the masks that constituted our labels. For these, parameters of the algorithm were chosen in a cohort-specific manner based on what was deemed optimal for each cohort. For the validation and evaluation sets of the tissue detection algorithm, the parameters were instead tuned to every individual WSI, and re-tuned until the visually evaluated quality of the mask was very high. Finally, for the comparison between thresholding and AI-based tissue detection, a single uniform set of parameters was used for all WSIs, selected manually based on what empirical testing revealed to be the most consistently good parameters during the validation set curation.

## Segmentation AI model

U-Net is a convolution neural network developed for segmentation of biomedical images,[40] and UNet++ is an extension of this architecture.[38] They are supervised learning models and hence require training data with annotated labels: WSIs or image patches with corresponding segmentation masks. The AI model architecture in this paper was that of UNet++, implemented using the SegmentationModels python library, version 0.3.3.[41] The implementation used encoder resnext101_32x4d with a depth of 5, and five decoder channels (512, 256, 128, 54, 32).

Three groups of augmentations from the Albumentations Python library (version 1.3.1) were used. Each group had a 50% probability to be applied, and within the groups the augmentations were applied with probability p specified below:

**Basic augmentations (p=0.5):**

- Vertical flip: albumentations.VerticalFlip (probability p=0.5)
- Horizontal flip: albumentations.HorizontalFlip (p=0.5)
- Random 90° rotations: albumentations.RandomRotate90 (p=1.0)

**Advanced augmentations (p=0.5):**

- Unsharp masking: albumentations.UnsharpMask (p=0.5)
  - blur_limit: [1, 51], alpha: [0.5, 1.0]
- Gaussian blurring: albumentations.GaussianBlur (p=0.5)
  - blur_limit: [1, 9]

- Color jitter: albumentations.ColorJitter (p=0.5)
    - brightness: [0.8, 1.2], contrast: [0.5, 1.5], saturation: [0.5, 1.5], hue: [-0.05, 0.05]
- Gamma correction: albumentations.RandomGamma (p=0.5)
    - gamma_limit: [80, 120]
- Random tone curve adjustment: albumentations.RandomToneCurve (p=0.5)
    - scale: 0.2

**Noise augmentations (p=0.5):**

- Gaussian noise: albumentations.GaussNoise (p=0.5)
    - var_limit: [1, 50]
- Multiplicative noise: albumentations.MultiplicativeNoise (p=0.5)
    - multiplier: [0.95, 1.05], element-wise
- Camera sensor noise: albumentations.ISONoise (p=0.5)
    - color_shift: [0.01, 0.05], intensity: [0.1, 0.5]
- Image compression artifacts: albumentations.ImageCompression (p=0.5)
    - quality_lower: 70

Training was done with an AdamW optimizer[42] (torch.optim.AdamW) with a base learning rate of 1e-6, epsilon constant of 1e-6 for stability, and weight decay 0.01. Binary cross entropy (torch.nn.BCEWithLogitsLoss) was used as a loss function for training, while F1-score (torchmetrics.classification.BinaryF1Score) was used as a metric on the validation set for early stopping.

# Tissue detection and patch extraction

For both the thresholding algorithm and segmentation AI, a resolution of 8.0 μm per pixel was used, which is heavily downsampled from the original resolution of the WSIs, and segmentation masks were stored as binary images. In training the segmentation AI, patches of size 512x512 pixels were extracted with no overlap. To fit an exact number of such patches, each WSI was first mirrored around each edge an appropriate amount. During inference, patches were generated such that they overlap 128 pixels about each edge to allow the edges of the predicted mask to be discarded. This avoids issues near tile edges due to lack of neighboring pixels providing context.

For both tasks, patches were downsampled from the closest higher resolution level in the WSI resolution pyramid using Lanczos resampling. For training the grading model, a higher resolution of 1.0 μm per pixel was used, with patches of size 256×256 pixels. Only patches with at least 10% of tissue pixels according to the segmentation masks were kept. Patches were extracted without overlap for training and with 128 pixel overlap during inference. Extracted patches were stored in TFRecord format, with each WSI saved as a separate file.

## Gleason grading model

The grading model used for evaluation in this study is a weakly-supervised algorithm relying on an attention-based multiple instance learning (ABMIL) architecture.[43] The model utilizes an EfficientNet-V2-S encoder[44] initialized with ImageNet weights that produces patch-level feature embeddings. These are then aggregated into slide-level representations through the ABMIL and classified into primary and secondary Gleason patterns (i.e. 3, 4, or 5), and further translated into Gleason scores and International Society of Urological Pathology (ISUP) grades. The model was trained in an end-to-end fashion, jointly optimizing all model parameters for cross-entropy loss using the AdamW optimizer with a base learning rate of 0.0001. Details on the model design, hyperparameters, and complete training strategy are given in the original publication.[35] The model was trained on 10 cross-validation folds, stratified by patient and ISUP grade. During model predictions, test time augmentation (TTA) was applied on three iterations for each of the 10 folds, and the final predictions were obtained as a majority vote of the resulting 30 Gleason scores.

## Statistical analysis

The segmentation masks produced by thresholding and AI were compared using the pixel-wise metrics of sensitivity (true positive rate) and precision (positive predictive value). For our purposes, a model with high sensitivity is crucial, as low sensitivity indicates large missed regions. Precision is included to ensure that excessive amounts of background are not detected as tissue.

The Gleason grading model was trained using pre-existing segmentation masks generated with thresholding, and evaluated once using the UNet++ segmentation masks and once using masks generated with thresholding to detect tissue in the evaluation slides. In this step, all thresholding masks were created using a uniform set of thresholding parameters, generated via empirical testing. The Gleason grading models were compared using quadratic weighted Kappa, a modification of the Cohen's Kappa statistic that measures agreement between two sets: in this case, the model's predictions and the pathologists' labels for each WSI or group of WSIs graded together. Confidence intervals were computed using bootstrapping with 1000 replicates.

# Results

## Tissue detection quality: thresholding vs AI

For measuring tissue detection quality, we calculated pixel-level sensitivity and precision between the evaluation set masks and the AI and thresholding segmentation masks, respectively. The sensitivity was of highest importance, since a low sensitivity indicates that the tissue detection has mistakenly categorized tissue as background. Low precision instead indicates that large amounts of background have been categorized as tissue.

The AI achieved an average sensitivity of 0.9840 (95% CI: 0.9833, 0.9848) and a precision of 0.9461 (95% CI: 0.9452, 0.9469) against the curated evaluation set masks, whereas the thresholding algorithm achieved 0.9804 (95% CI: 0.9790, 0.9819) and 0.9650 (95% CI: 0.9641, 0.9658), respectively. Two WSIs were excluded from calculating precision due to not having any detected tissue. The distribution of sensitivity and precision for each cohort of the evaluation set can be seen in Figure 1. Both algorithms performed highly but, importantly, the thresholding algorithm failed drastically in terms of sensitivity on a small set of individual WSIs, while the AI achieved more acceptable worst case performance.

## Does tissue detection influence downstream tasks?

The model trained using thresholding-based tissue detection masks was used to run predictions on the internal and external validation cohorts, using either thresholding- or AI-based tissue detection on these validation data. In the cases where one or both tissue detection models failed to detect tissue completely, the WSIs were discarded from the analysis. The number of WSIs for which this occurred is shown in Table 2. For all WSIs where both tissue detection models identified any tissue, the quadratic kappa statistics quantifying the concordance between the ISUP grades predicted by the model and reported by the pathologists for each cohort are displayed in Figure 2. No cohorts had non-overlapping confidence intervals, implying there is no large difference in overall performance between the two models. In Supplementary Figure 1, a four-way comparison is shown which additionally includes a model trained on AI-based tissue detection masks.

## Tissue detection -dependent variations in AI Gleason grading

We identified all the slides with per-slide pathologist grading available, where the two tissue detection algorithms led to different predictions of ISUP grade. This occurred for 163 slides out of 11,350 (1.4%), or 120 out of 3,459 (3.5%) malignant slides. Four example WSIs with the true label ISUP 2 or higher were chosen for visualization in Figure 3. In each case, it is clear why the choice of tissue detection mask was important for making the correct prediction. This elucidates what may go wrong when using automated tissue detection. In a clinical setting, unless the system flags for potentially faulty tissue detection (which may be algorithmically challenging) or the tissue detection can be easily visualized by a pathologist, none of these cases would likely be detected as erroneous.

# Discussion

The sensitivity distribution for the tissue detection models in Figure 1 indicates that they both aptly caught tissue in the vast majority of cases, but the thresholding model missed large areas of tissue more often than the AI model. Since the cases where large pieces of tissue are missing are the most likely to result in failed cancer detection, improving the sensitivity of tissue detection was the most important task. To this end, the tissue detection AI succeeded in outperforming thresholding. The higher precision of the thresholding algorithm can be explained in part by the fact that the masks in the evaluation set also were generated via thresholding, and hence share some pixel-level characteristics that the tissue detection AI may stray from. The results in Figure 2 indicate that no statistically significant overall difference in ISUP grade predictions could be observed between the two tissue detection approaches, but the existence of outlier cases where the tissue detection was highly important is evident from Table 2 and Figure 3.

While there is no consensus on the ideal way to do automatic tissue detection, we contend that similarly to other image analysis tasks, a well-trained AI is likely to be more reliable than classical methods, especially for samples that stray from the typical appearance of these images. To that end, one potential improvement of our model could be to train using more color augmentations to avoid issues where the AI overlearns color associations, which would make it more robust to unusual color profiles, such as the pale white tissue of Figure 3b.

Many digital pathology projects utilize tissue detection as a computationally cheap way to remove background from analyses. Thish both saves compute time and physical energy resources, as well as reduces background-related problems such as shortcut learning[45] due to pathologists' pen markings. For these reasons, tissue detection is likely to remain relevant even for slide-level analytics and the digital pathology community should ensure the process is reliable, reproducible, and consistent.

There is an argument to be made regarding foundation models. Models such as UNI[46] and Virchow[47] are trained without preprocessing normalisation procedures in order for the training set to be as diverse as possible. However, these models utilize simple tissue detection methods relying on thresholding and a simple hue-based detection, respectively. If these methods consistently fail to detect tissue outside allowed ranges, the training philosophy of allowing diverse training data to improve robustness is at least in part compromised. Future papers may examine the effect that different tissue detection methods have when utilising foundation models as feature extractors, and whether there is a drop in performance for tissue that is not captured by the tissue detection models used in their training.

A limitation of the study is the lack of a standardized procedure for obtaining training data for the AI segmentation model. Naturally, we wanted to train on the highest quality segmentation masks we had available, which were not generated in a reproducible manner. We believe, however, that a high-performing U-Net type model can be trained without any complicated procedures for generating training labels, for instance by training it directly on the labels generated via a thresholding algorithm without manual refinement. In our own data, only 0.16% of masks had been manually refined, and the cohort-specific thresholding parameters did not considerably outperform the uniform parameters we used in this paper. This indicates that weak labels can be sufficient for training an AI segmentation model at least for relatively simple tasks like tissue detection. Furthermore, research groups utilising tissue segmentation will already have segmentation masks available to them, and those masks can constitute their training labels.

A strength of this study is the large amount of data available. However, it is noteworthy that tissue detection can be learned with smaller models utilising less data as well, and we do not contend that training data amounts similar to the ones used in this study are necessary to incorporate AI tissue detection into a digital pathology pipeline. Since it is in the interest of the community to make models smaller and more efficient both to improve accessibility to clinics and to reduce carbon emissions,[48] future studies can examine training smaller tissue detection models.

In this paper, we have shown that choice of tissue detection method can influence how often the diagnostic AI process malfunctions by failing to detect any tissue in a slide. In this respect, an AI tissue detection method was more reliable than a classical thresholding algorithm. The AI method also successfully detected a larger portion of tissue when the segmentation masks were evaluated directly. When evaluating downstream effects on a diagnostic Gleason grading algorithm, we observed that the model's predictions were influenced by the tissue detection method in a clinically non-negligible number of slides, despite the difference in overall performance not being statistically significant. As clinical application of these systems has become reality, it is increasingly important that every part of these algorithms works accurately and consistently to ensure the efficiency and patient safety of diagnostic AI in all situations.

# Ethical considerations

This study included data gathered in one or more collection rounds at participating international sites between 2012 and 2024. All datasets were de-identified at their respective sites and subsequently transferred to Karolinska Institutet in an anonymized format. This study complies with the Helsinki Declaration. The patient sample collection was approved by the Stockholm Regional Ethics Committee (permits 2012/572-31/1, 2012/438-31/3, and 2018/845-32), the Swedish Ethical Review Authority (permit 2019-05220), and the Regional Committee for Medical and Health Research Ethics in Western Norway (permits REC/Vest 80924, REK 2017/71). Informed consent was obtained from patients in the Swedish dataset and was waived for other data cohorts due to the use of de-identified prostate specimens in a retrospective setting. Patient involvement in this study was supported by the Swedish Prostate Cancer Society.

# Acknowledgments

A.B. received a grant from the Health Faculty at the University of Stavanger, Norway. B.G.P and K.D.S received funding from Innovation Fund Denmark and Nordforsk (Grant no. 8114-00014B) for the Danish branch of the NordCaP project. K.H. received funding from Business Finland (BF) for the Finnish branch of the NordCaP project. K.K. received funding from the SciLifeLab & Wallenberg Data Driven Life Science Program (KAW 2024.0159), David and Astrid Hägelen Foundation, Instrumentarium Science Foundation, KAUTE Foundation, Karolinska Institute Research Foundation, Orion Research Foundation and Oskar Huttunen Foundation.

We want to thank Carin Cavalli-Björkman, Astrid Björklund and Britt-Marie Hune for assistance with scanning and database support. We would also like to thank Simone Weiss for assistance with scanning in Aarhus, and Silja Kavlie Fykse and Desmond Mfua Abono for scanning in Stavanger. We would like to acknowledge the patients who participated in the STHLM3 diagnostic study and the OncoWatch and NordCaP projects and contributed the clinical information that made this study possible. Computations were enabled by the National Academic Infrastructure for Supercomputing in Sweden (NAISS) and the Swedish National Infrastructure for Computing (SNIC) at C3SE partially funded by the Swedish Research Council through grant agreement no. 2022-06725 and no. 2018-05973, and by the supercomputing resource Berzelius provided by the National Supercomputer Centre at Linköping University and the Knut and Alice Wallenberg Foundation.

# Competing interests

N.M., L.E., and K.K. are shareholders of Clinsight AB.

# Data availability

A subset of the data used for development (STHLM3 and RUMC cohorts) is available for non-commercial purposes subject to a CC BY-SA-NC 4.0 license as part of the PANDA challenge dataset and is freely downloadable after registration at https://www.kaggle.com/c/prostate-cancer-grade-assessment. This study also utilizes publicly available datasets used for external validation (SPROB20, UKK, and WNS). The SPROB20 cohort is available for non-commercial purposes under the AIDA BY license upon accepted access request at the AIDA Data Hub at https://datahub.aida.scilifelab.se/10.23698/aida/sprob20. UKK and WNS cohorts are available for non-commercial purposes under the CC BY-NC-SA 4.0 license upon accepted access request at https://zenodo.org/records/8102833 and https://zenodo.org/records/8102929. Data from additional sources cannot be shared publicly. For any requests to access these sources, inquiries should be directed to K.K. at Karolinska Institutet. Requests will be evaluated on a case-by-case basis, with approvals granted if they comply with data privacy regulations and intellectual property policies.

# Code availability

Core components of the segmentation and grading models relied on open-source repositories. For the grading model we used [https://github.com/AMLab-Amsterdam/AttentionDeepMIL](https://github.com/AMLab-Amsterdam/AttentionDeepMIL) and [https://github.com/huggingface/pytorch-image-models](https://github.com/huggingface/pytorch-image-models), and for the segmentation model we used [https://github.com/qubvel-org/segmentation_models.pytorch](https://github.com/qubvel-org/segmentation_models.pytorch). Torch ([https://github.com/pytorch/pytorch](https://github.com/pytorch/pytorch)) was used for model training and prediction. Scipy ([https://github.com/scipy/scipy](https://github.com/scipy/scipy)) and scikit-image ([https://github.com/scikit-image/scikit-image](https://github.com/scikit-image/scikit-image)) were used for the thresholding algorithm. Albumentations ([https://github.com/albumentations-team/albumentations](https://github.com/albumentations-team/albumentations)) was used for applying augmentations during model training. All steps involved in model training and design have been thoroughly documented in the Methods section to allow independent replication.

# Author contributions

S.E.B. developed the segmentation model and conducted statistical analyses. N.M., X.J. and K.K. developed the Gleason grading model. S.E.B., N.M., X.J., K.S. and K.K. all developed significant parts of the code used in this project and pre-processed datasets. A.B., E.G., E.A.M.J., S.R.K., J.A., M.G., P.L., M.B., R.K., R.Ł., K.H., P.V., B.G.P., K.D.S., B.P.U. and L.E. collected, curated and contributed clinical datasets. K.K. supervised the study. S.E.B., N.M. and K.K. drafted the manuscript. All authors read and approved the manuscript.

a modelling study. *Lancet Digit. Health* **6**, e58–e69 (2024).

# Figures and Tables

| Split | Cohort | Number of patients | Scanner | Number of WSIs | Label strength | Number of manually edited labels |
|---|---|---|---|---|---|---|
| Training | Stockholm3 | 2,444 | Aperio | 2,196 | Strong | 34 (1.5%) |
| | | | Hamamatsu | 3,976 | Strong | 10 (0.25%) |
| | | | Philips | 23,501 | Weak | 0 (0%) |
| | Stavanger University Hospital | 639 | Hamamatsu | 4,150 | Weak | 0 (0%) |
| Validation | Stockholm3 | 116 | Aperio | 131 | Strong | N/A |
| | | | Hamamatsu | 874 | | |
| | | | Philips | 1,146 | | |
| | Stavanger University Hospital | 30 | Hamamatsu | 201 | | |
| | Radboud University Medical Center | 33 | 3DHISTECH | 142 | | |
| | Capio S:t Göran Hospital | 5 | Aperio | 25 | | |
| | | | Hamamatsu | 4 | | |
| Evaluation | Stockholm3 | 179 | Aperio | 225 | | |
| | | | Hamamatsu | 1314 | | |
| | | | Philips | 1661 | | |
| | Stavanger University Hospital | 41 | Hamamatsu | 254 | | |
| | Radboud University Medical Center | 137 | 3DHISTECH | 516 | | |
| | Karolinska University Hospital | 73 | Hamamatsu | 330 | | |

**Table 1:** Summary of data and partitions for training, validation and evaluation for development of the segmentation model. The labels were considered strong if tissue segmentation masks from these cohorts had been checked visually to verify their correctness, and weak if the quality of almost all segmentation masks in the cohort had not been checked. Certain cohorts that had been checked visually had also had a small number of segmentation masks manually edited to improve the labels, the number of which is documented in the rightmost column.

| Cohort | Tissue detection failure by AI only (n WSIs) | Tissue detection failure by thresholding only (n WSIs) | Tissue detection failure by both AI and thresholding (n WSIs) | Total slides |
|---|---|---|---|---|
| Aarhus University Hospital | 0 | 0 | 0 | 102 |
| Hospital Wiener Neustadt | 0 | 0 | 0 | 50 |
| Medical University of Lodz | 0 | 79 | 8 | 2,435 |
| Mehiläinen Länsi-Pohja* | 0 | 0 | 0 | 1,963 |
| Radboud University Medical Center | 0 | 0 | 0 | 516 |
| Spear Prostate Biopsy 2020** | 0 | 8 | 0 | 2,570 |
| Stavanger University Hospital | 0 | 1 | 1 | 1,199 |
| Stockholm3 | 0 | 6 | 1 | 14,907 |
| Synlab Finland* | 0 | 2 | 0 | 536 |
| Synlab France | 0 | 2 | 0 | 515 |
| Synlab Switzerland* | 2 | 18 | 12 | 2,429 |
| University Hospital Cologne | 0 | 0 | 0 | 50 |
| **Total** | 2 (<0.01%) | 116 (0.43%) | 22 (0.08%) | 27,272 |

*These cohorts were evaluated on location level

**This cohort was evaluated on patient level

**Table 2:** The number of WSIs in the test cohorts of the downstream Gleason grading model where tissue segmentation by AI, by thresholding or by both methods failed to detect any tissue. Only WSIs with tissue detected by both algorithms were included in the comparison (**Figure 2**).

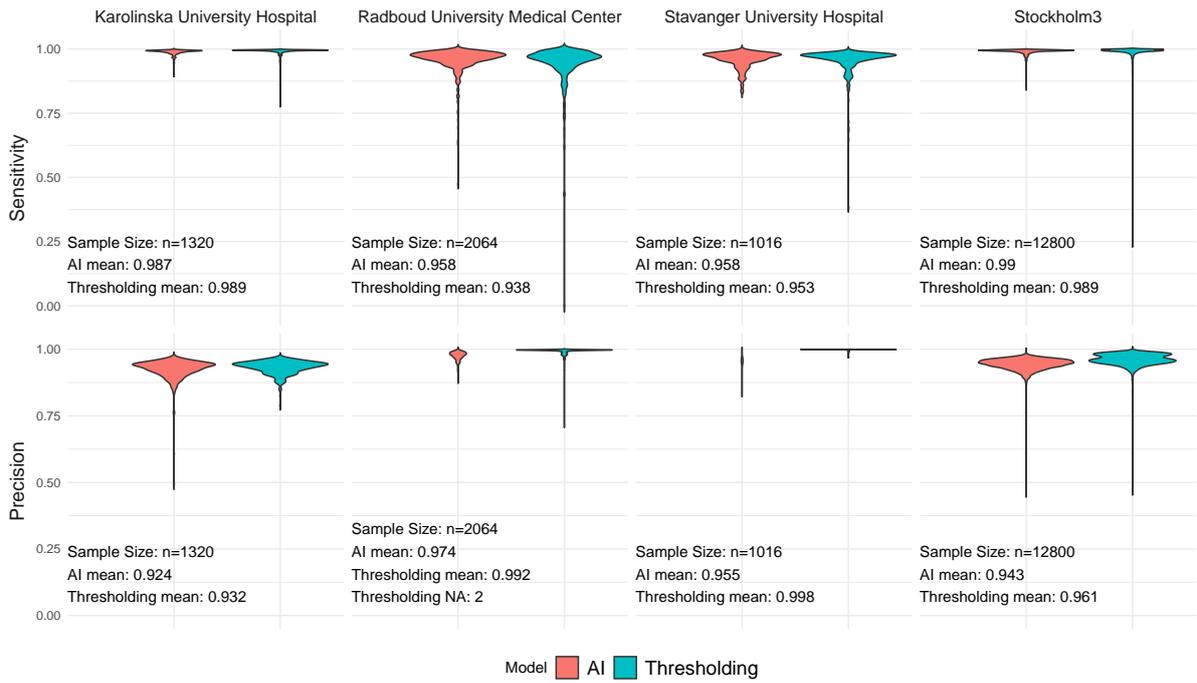

**Figure 1:** A violin plot of the pixel-wise sensitivity (top row) and precision (bottom row) of the tissue detection AI and the classical thresholding-based tissue detection. The ground truth tissue/background labels are curated high quality segmentation masks generated with thresholding.

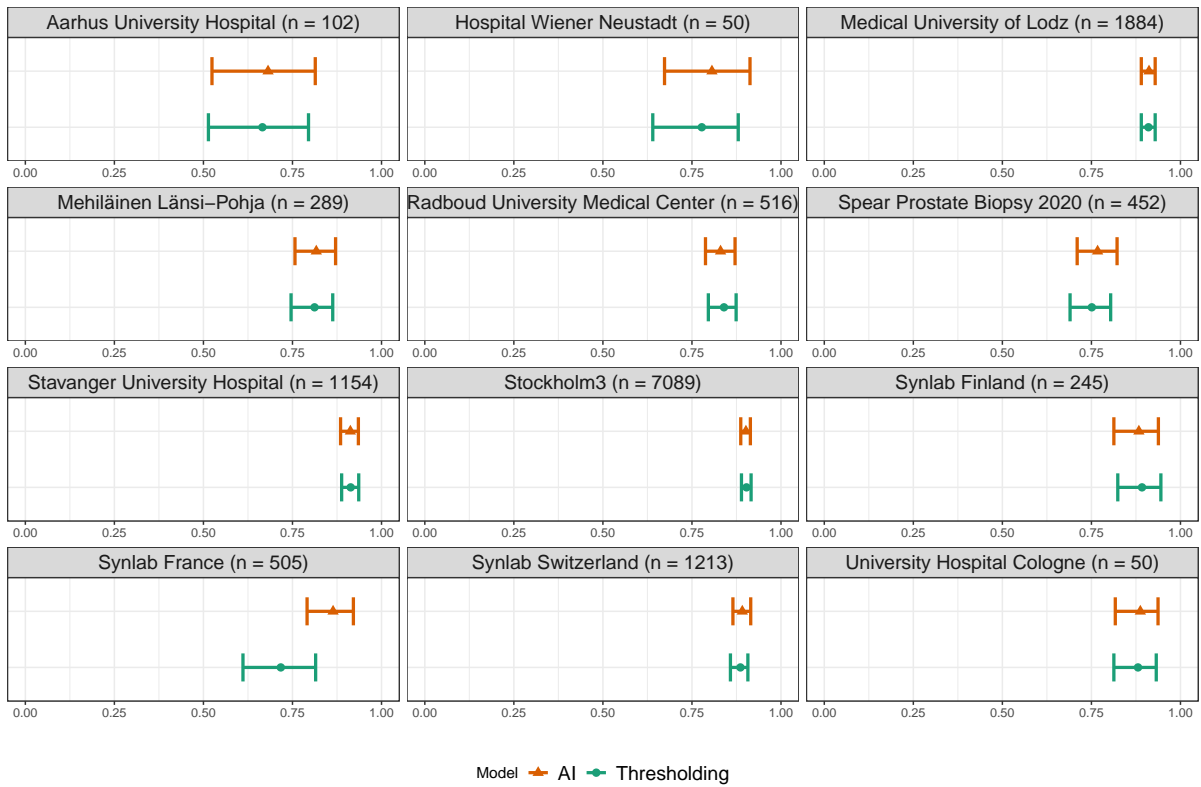

**Figure 2**: The concordance between the predictions of ISUP grade by the Gleason grading model and the reference grading by pathologists, measured by Cohen's quadratically weighted kappa statistic. Results are shown based on tissue detection on the validation slides using AI-based and thresholding-based methods. The dots indicate point estimates on the entire dataset and the whiskers indicate 95% CIs. Confidence intervals were estimated with bootstrapping using 1000 replicates. Only cases where both models were able to detect any tissue were included, see **Table 2**. Synlab Finland, Synlab Switzerland and Mehiläinen Länsi-Pohja had reference grading per anatomical location and SPROB had reference grading per patient; WSIs from these cohorts were pooled to get predictions at location and patient levels, respectively.

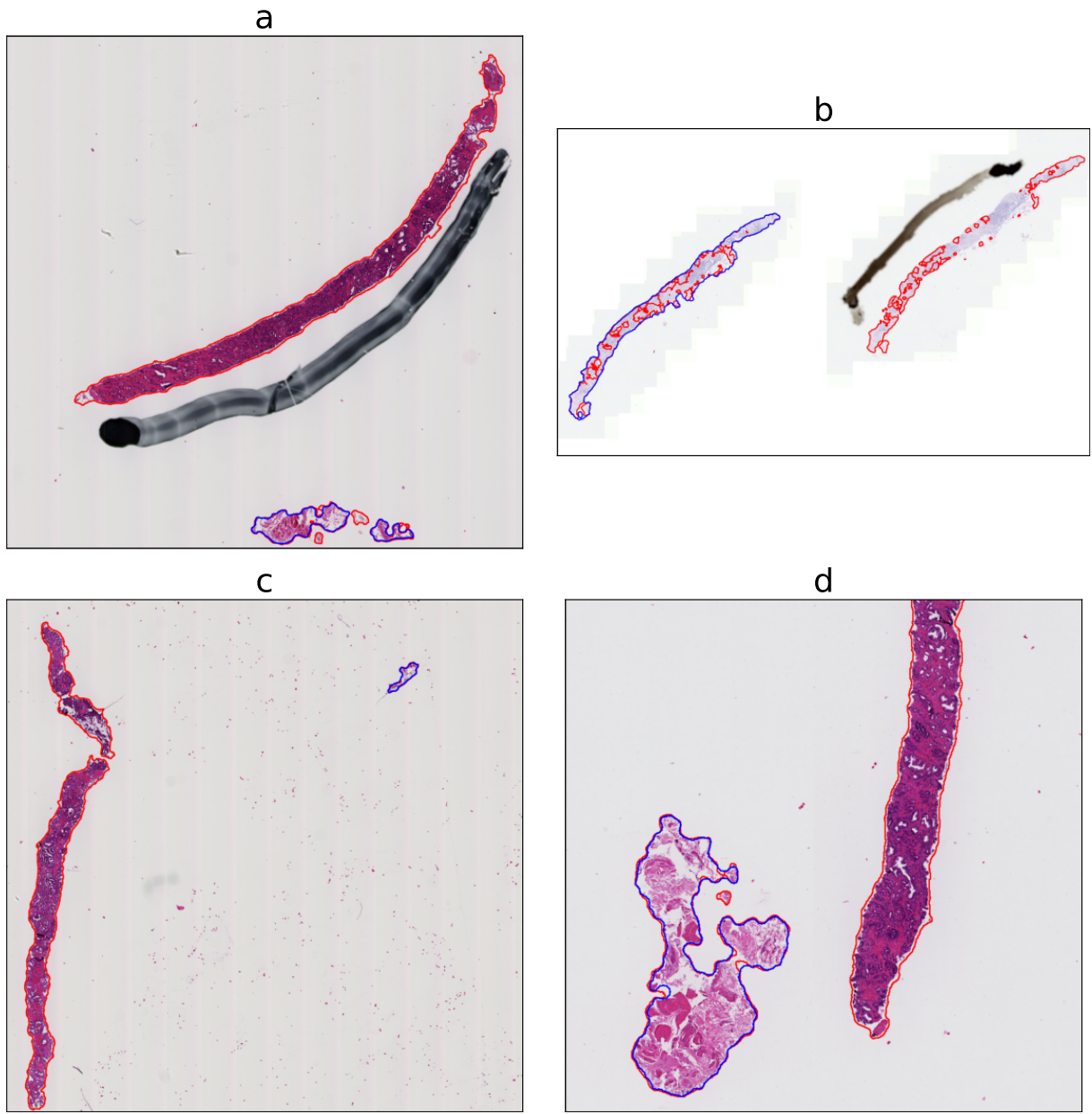

**Figure 3:** Outlines of tissue segmentation masks for AI (red) and thresholding (blue) in example cases where cancer grading by the downstream model was affected by differences in tissue detection. **(a):** The thresholding model missed an entire large piece of tissue and subsequently the Gleason grading model predicted ISUP 0 when it is in fact ISUP 2. **(b):** Both methods struggled to properly segment tissue on this slide: the thresholding model missed an entire piece of tissue while the AI model only detected some chunks. The Gleason grading AI is still able to predict ISUP 3, the correct grade, using the thresholding segmentation mask but predicts ISUP 2 using the AI segmentation mask. **(c):** Both models incorrectly segment some debris, but only the AI model detects the large piece of tissue. Subsequently, the Gleason grading algorithm predicts ISUP 0 using the thresholding mask and ISUP 3, the correct grade, using the AI mask. **(d):** Zoomed in for clarity. Both models segment a piece of tissue at the bottom but the thresholding segmentation completely fails to identify a much larger piece of tissue extending high

above the edge of the cropped image. Subsequently, the Gleason grading AI predicts ISUP 0 when using the thresholding-based segmentation mask and ISUP 2, the correct grade, when using the AI mask.

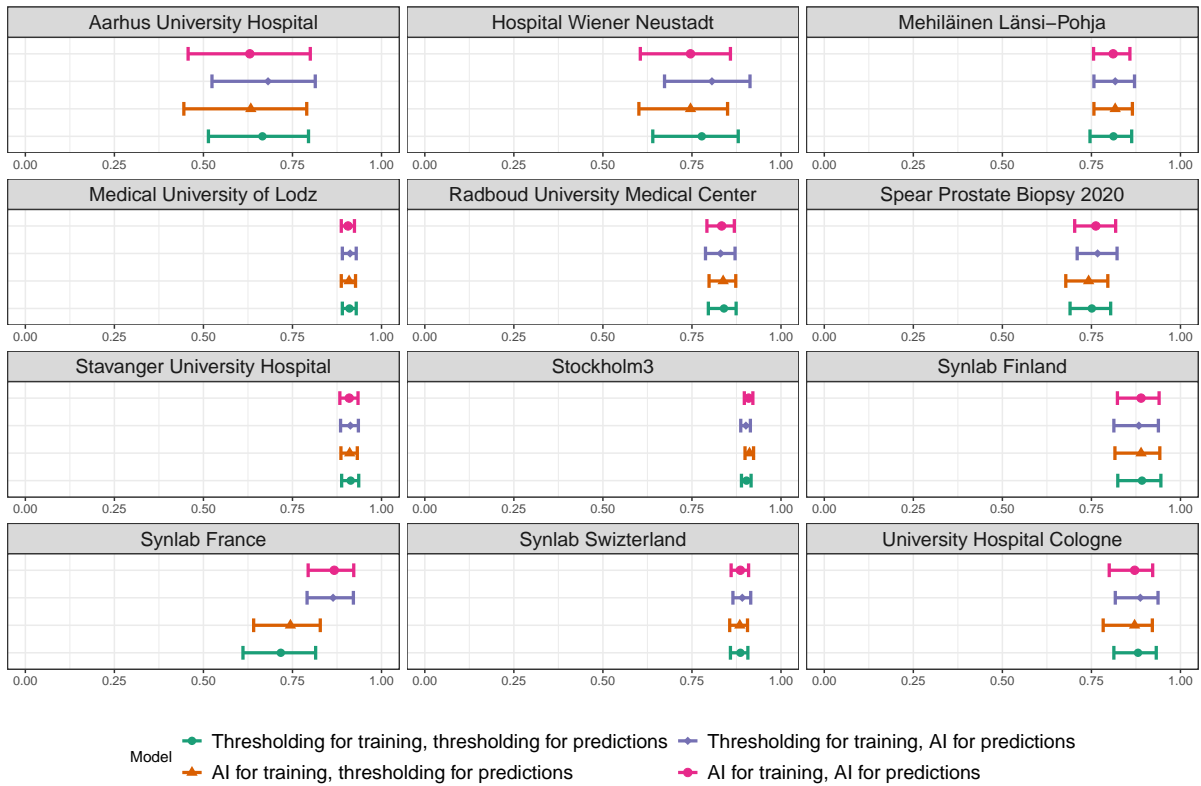

**Supplementary Figure 1:** The concordance between the predictions of ISUP grade by two versions of the Gleason grading model and the reference grading by pathologists, measured by Cohen's quadratically weighted kappa statistic. Results are shown based on tissue detection on the validation slides using AI-based and thresholding-based methods. One of the Gleason grading models was trained using tissue segmentation masks that had been generated using thresholding ("thresholding model") and one was trained using tissue segmentation masks that had been generated using the tissue detection AI ("AI model"), and each was evaluated using segmentation masks generated with each of the methods ("thresholding tiles", "AI tiles"). The dots indicate point estimates on the entire dataset and the whiskers indicate 95% CIs. Confidence intervals were estimated with bootstrapping using 1000 replicates. Only cases where both models were able to detect any tissue were included, see **Table 2**. Synlab Finland and Synlab Switzerland had reference grading per anatomical location and SPROB had reference grading per patient; WSIs from these cohorts were pooled to get predictions at location and patient levels, respectively.